\documentclass[10pt,twocolumn,letterpaper]{article}

\usepackage[pagenumbers]{cvpr}      
\usepackage{multirow}

\definecolor{cvprblue}{rgb}{0.21,0.49,0.74}
\usepackage[pagebackref,breaklinks,colorlinks,allcolors=cvprblue]{hyperref}
\usepackage{colortbl}  
\usepackage{xcolor}
\usepackage[utf8]{inputenc}
\usepackage{url}
\usepackage{booktabs}
\usepackage{makecell}
\usepackage{amssymb}
\usepackage{bbding}
\usepackage{pifont}
\usepackage{wasysym}
\usepackage{utfsym}
\usepackage{fontawesome}
\usepackage{algorithm}
\usepackage{algorithmicx}
\usepackage{algpseudocode}
\definecolor{cvprblue}{rgb}{0.21,0.49,0.74}
\usepackage[pagebackref,breaklinks,colorlinks,allcolors=cvprblue]{hyperref}

\def\confName{CVPR}
\def\confYear{2026}

\title{On-the-fly Large-scale 3D Reconstruction from Multi-Camera Rigs}

\author{
  Yijia Guo$^{1*}$ \quad
  Tong Hu$^{2*}$ \quad
  Zhiwei Li$^3$ \quad
  Liwen Hu$^{1}$ \quad
  Keming Qian$^2$ \\
  Xitong Lin$^{4}$ \quad
  Shengbo Chen$^{3}$ $^{\text{✉}}$ \quad
  Tiejun Huang$^{1}$  \quad
  Lei Ma$^{1,2}$ $^{\text{✉}}$\\
  $^1$ \small State Key Laboratory of Multimedia Information Processing, School of Computer Science, Peking University \\
  $^2$ \small National Biomedical Imaging Center, Peking University \\ 
 $^3$ \small School of Software, Nanchang University \\
 $^4$ \small Shenzhen International Graduate School, Tsinghua University
}

\let\oldtwocolumn\twocolumn
\renewcommand\twocolumn[1][]{%
    \oldtwocolumn[{#1}{
    \vspace{-1.1cm}
    \begin{center}
           \includegraphics[width=1.9\columnwidth]{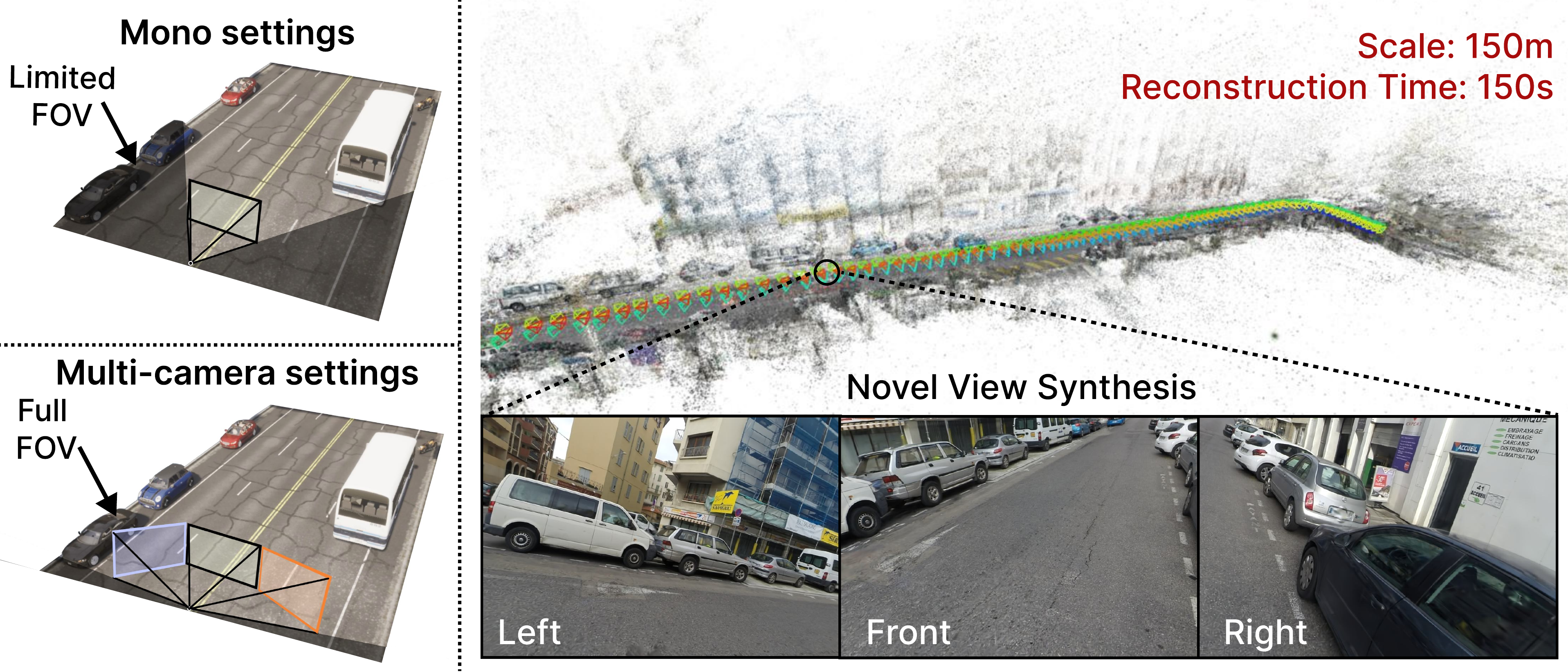}
            \vspace{-0.2cm}
           \captionof{figure}{\textbf{Left:} Comparision of mono and multi-camera settings. The multi-camera rig provides more complete coverage of the scene. \textbf{Right:} Our approach enables on-the-fly and comprehensive 3D reconstruction of large-scale scenes from multi-camera rig streams.}
           \vspace{-0.0cm}
           \label{fig1}
        \end{center}
    }]
}

\newcommand{\thickhline}{%
    \noalign {\ifnum 0=`}\fi \hrule height 1pt
    \futurelet \reserved@a \@xhline
}

\begin{document}
\maketitle
\renewcommand*{\thefootnote}{$*$}\footnotetext{Equal contributions.\hspace{1em}%
$^{\text{✉}}$Corresponding authors.}
\renewcommand*{\thefootnote}{$\ddagger$}
\footnotetext{This work was supported by National Science and Technology Major Project (2022ZD0116305).}
\begin{abstract}
Recent advances in 3D Gaussian Splatting (3DGS) have enabled efficient free-viewpoint rendering and photorealistic scene reconstruction. While on-the-fly extensions of 3DGS have shown promise for real-time reconstruction from monocular RGB streams, they often fail to achieve complete 3D coverage due to the limited field of view (FOV). Employing a multi-camera rig fundamentally addresses this limitation.
In this paper, we present the first on-the-fly 3D reconstruction framework for multi-camera rigs. Our method incrementally fuses dense RGB streams from multiple overlapping cameras into a unified Gaussian representation, achieving drift-free trajectory estimation and efficient online reconstruction. We propose a hierarchical camera initialization scheme that enables coarse inter-camera alignment without calibration, followed by a lightweight multi-camera bundle adjustment that stabilizes trajectories while maintaining real-time performance. Furthermore, we introduce a redundancy-free Gaussian sampling strategy and a frequency-aware optimization scheduler to reduce the number of Gaussian primitives and the required optimization iterations, thereby maintaining both efficiency and reconstruction fidelity.
Our method reconstructs hundreds of meters of 3D scenes within just 2 minutes using only raw multi-camera video streams, demonstrating unprecedented speed, robustness, and Fidelity for on-the-fly 3D scene reconstruction.
\end{abstract}    
\section{Introduction}

Recent advances in neural radiance field representations, such as 3D Gaussian Splatting (3DGS) \cite{3dgs}, have enabled highly efficient and photorealistic 3D reconstruction from casually captured images, supporting free-viewpoint navigation and real-time rendering. Building upon these advances, a growing number of works have integrated 3DGS into SLAM \cite{matsuki2024gaussian, zheng2025wildgs} or feed-forward reconstruction frameworks \cite{chen2024mvsplat,liu2024mvsgaussian,zhang2024gslrm,tang2024hisplat}, enabling on-the-fly 3D scene reconstruction directly from streaming RGB inputs. These methods have shown great potential in extended reality (XR), robotics, and autonomous driving, and are increasingly applied to large-scale outdoor environments, further broadening their impact.

However, most existing on-the-fly 3DGS-based reconstruction methods are limited to monocular video inputs, resulting in incomplete sampling of the scene and missing cross-view consistency, as shown in Fig.~\ref{fig1}. A straightforward and effective way to address this limitation is to employ synchronized multi-camera rigs, which can capture overlapping views of the environment and thus provide more complete spatial coverage. This design is also adopted by most publicly available large-scale outdoor datasets \cite{li2023matrixcity,sun2020scalability,zurn2024wayvescenes101}, especially those for road-scene reconstruction, as it improves coverage efficiency, reduces capture time, and enhances overall data quality.
Although capturing data with a multi-camera rig is straightforward, achieving real-time, large-scale multi-camera reconstruction remains a formidable challenge. Existing SLAM-based methods for multi-camera systems typically depend on inertial sensors, precise extrinsic calibration, or offline optimization, which limits their scalability and adaptability in unconstrained outdoor settings. Furthermore, on-the-fly reconstruction in this regime must simultaneously meet several demanding requirements: a) Updating large-scale Gaussian scenes within strict time limits (typically under 1 second per keyframe) while preserving visual fidelity. b) Ensuring inter-camera scale consistency and stable trajectories across wide baselines. c) Avoiding uncontrolled growth in the number of Gaussian primitives.


We present the first on-the-fly 3D reconstruction framework specifically designed for large-scale scenes captured by multi-camera rigs. Our method completely reconstructed \textbf{100$\mathbf{m}$ street scenes or 100,000$\mathbf{m^2}$ aerial scenes in two minutes, without requiring explicit camera calibration.} To establish reliable inter-camera alignment, we propose a initialization strategy that automatically identifies a central reference camera via feature-based matching and progressively aligns other cameras in a tree-structured manner. This initialization provides a coarse yet globally consistent alignment for subsequent optimization. On top of this, we introduce a lightweight multi-camera bundle adjustment (BA) scheme that maintains efficiency while suppressing trajectory drift and avoiding local minima. Instead of relying on traditional density-boosting or oversampling heuristics, we perform redundancy-free Gaussian sampling and merging across overlapping camera views, which accelerates convergence and reduces computational overhead. Finally, a frequency-aware optimization scheduler dynamically allocates more iterations to regions requiring rapid refinement, enabling fast and stable scene convergence with strong global consistency. We demonstrate state-of-the-art performance on large-scale outdoor reconstruction tasks from both street and aerial datasets. In addition, we construct a new dataset featuring synchronized multi-camera captures, which further validates the robustness of our approach. Our main contributions are summarized as follows:
\begin{itemize}
    \item The first on-the-fly 3D reconstruction framework for multi-camera rigs, capable of online reconstructing kilometer-scale outdoor scenes.
    \item A lightweight multi-camera bundle adjustment jointly with redundancy-free Gaussian sampling and frequency-aware optimization, boosting both novel View synthesis and pose estimation quality compared with the existing methods.
    \item A high-quality outdoor dataset for multi-camera reconstruction, covering over $1Km$ street scenes and 1$Km^2$ aerial scenes, with careful treatment of exposure inconsistency, desynchronization, and motion blur.
\end{itemize}

\section{Related Work}
\label{sec:relatedwork}

\textbf{On-the-fly 3D Reconstruction.} The integration of 3DGS and SLAM pipelines \cite{keetha2023splatam, yan2023gs, yugay2023gaussian, hu2024cg,matsuki2024gaussian} provides online tracking, mapping, and maintaining photorealistic reconstruction. But early methods suffered from error accumulation at scale. DroidSplat \cite{homeyer2025droid} achieves high-quality mapping through geometric priors, yet its offline two-stage design precludes real-time use. Recent efforts address memory efficiency \cite{deng2024compact} and optimization artifacts \cite{sun2024high}. While these methods successfully construct scenes from sensor streams in real-time, they struggled under wide baselines and large-scale outdoor environments.

To overcome the inherent limitations of SLAM-based approaches, which struggle to accommodate wide-baseline configurations and typically neglect high-fidelity rendering quality, researchers have increasingly turned to alternative methodological frameworks.
A parallel line of research explores feed-forward methods to directly infer point clouds or 3D Gaussians from images via network generalization.   Early approaches \cite{charatan2024pixelsplat, zou2024triplane, szymanowicz2024splatter, chen2024mvsplat, liu2024mvsgaussian, yang2025fast3r} establish one-shot but fast 3D reconstruction, while subsequent studies \cite{zhuo2025streaming,mahdi2025evict3r} rapidly extended this capability to video streams. However, these approaches still primarily focus on camera trajectory and geometric accuracy, paying limited attention to rendering quality and lacking scalability to large-scale scenes.
ARTDECO \cite{li2025artdeco} combines feed-forward networks with incremental Gaussian sampling to balance speed, scalability, and rendering quality; however, its reconstruction speed still falls short of on-the-fly requirements. Some feed-forward-free approaches have achieved this goal.
Gaussian On-the-Fly Splatting \cite{xu2025gotf} enhances robustness under pose uncertainty. On-the-fly NVS \cite{meuleman2025on} jointly estimates poses and reconstructs scenes via Gaussian sampling, demonstrating robustness in outdoor settings. Online-3DGS-Monocular \cite{wu2025monocular} enhances detail preservation in monocular modeling. However,all the aforementioned methods are limited to monocular input, making it difficult to reconstruct complete 3D scenes in many cases. Moreover, none of them can handle video streams captured by complex camera rigs. How to achieve real-time reconstruction of complete, large-scale 3D scenes from camera-rig video streams remains an open and challenging problem.

\textbf{3D Reconstruction for Camera Rigs:} Early SLAM systems for camera rigs \cite{ji2015particle, shi2012gps, kaess2010probabilistic, heng2015leveraging} include Multicol-SLAM \cite{urban2016multicol} and Pan-SLAM \cite{ji2020panoramic}, which provide solutions for rigid multi-fisheye setups. These methods typically require precise calibration and face challenges with wide baselines and dynamic environments. MCGS-SLAM \cite{cao2025mcgs} specifically targets complex multi-camera setups to enhance scale consistency and cross-view robustness.However, it typically rely on offline optimization to achieve high-quality reconstruction results. Rig3R \cite{li2025rig3r} is the first to employ a feed-forward network to process outputs from a multi-camera rig, enabling the estimation of both trajectories and point clouds. All these methods rely on camera calibration, lack validation on large-scale outdoor scenes, and offer limited rendering quality. In contrast, our approach simultaneously overcomes all of these limitations.
\section{Method}
\begin{figure*}[t]
  \centering
  \vspace{-0.5cm}
  \includegraphics[width=0.98\linewidth]{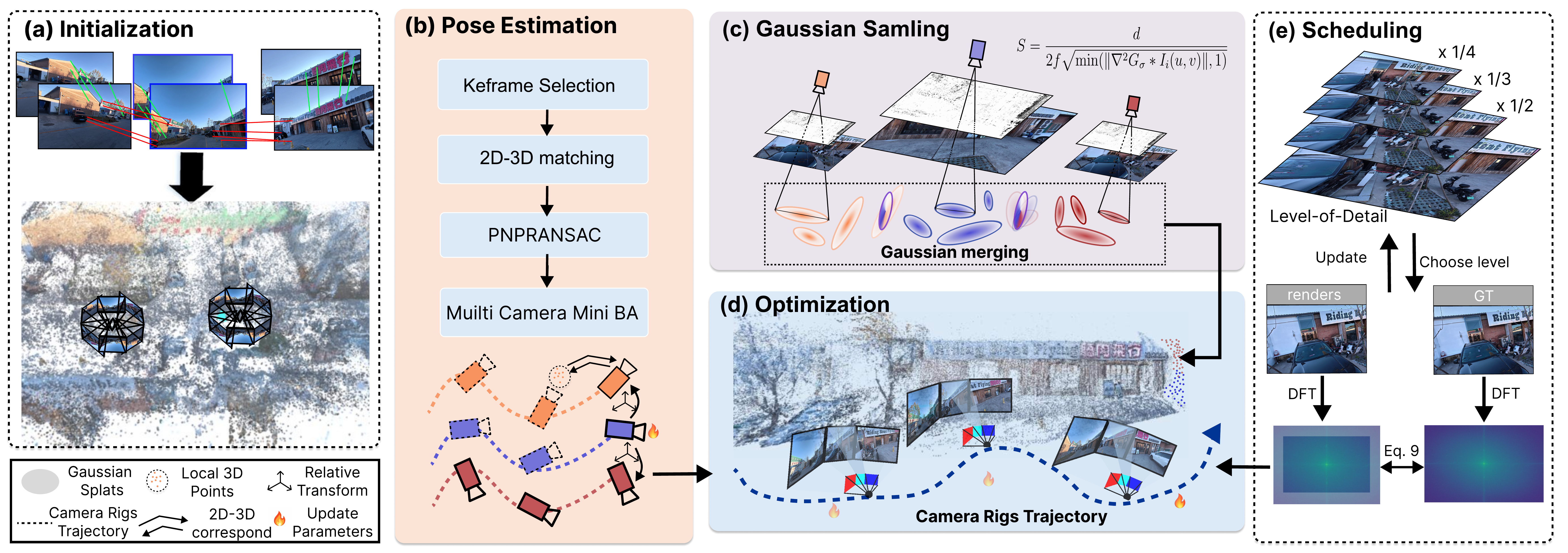}
  \caption{Overview of our on-the-fly pipeline.  We first initialize a central multi-camera rig system via feature matching and estimate the corresponding 3DGS scene and camera pose. Next, a lightweight multi-camera bundle adjustment refines trajectories across wide baselines for subsequent keyframes (Sec.~\ref{subsec_pose}). Instead of relying on traditional density-boosting or oversampling heuristics, we perform redundancy-free Gaussian sampling and merging across overlapping camera views, which accelerates convergence and reduces computational overhead. The sampled Gaussian primitives are progressively fused into the global scene during sequential frame updates (Sec.~\ref{subsec_Gaussian}). Finally, a frequency-aware optimization scheduler dynamically allocates more iterations to regions requiring rapid refinement, enabling fast and stable scene convergence with strong global consistency. (Sec.~\ref{subsec_optimize})}
  \label{fig_framework}
  \vspace{-0.5cm}
\end{figure*}
\textbf{Proble settings:}
Given a temporally ordered stream of images $\mathbb S_k= \{I^t_k\}_{t=0}^N$ captured at time t by camera $k$ in a camera rig consisting of $K$ cameras, our method is designed to simultaneously optimize the metric camera trajectory ${Traj}_k = \{P^t_k\}_{t=0}^N$ where $P^t_k$ is the camera pose for camera $k$ at time $t$ and Gaussian primitives $\mathbb G$ as
shown in , mathematically formulated as jointly:
\begin{align}
    \{ \{\mathbb S_k\}_{k=0}^K \} \rightarrow{}\{\mathbb{G}, \{{Traj}_k\}_{k=0}^K \}
\label{opti}
\end{align}
The multi-camera rig features overlapping fields of view and is rigidly linked and temporally synchronized during capture. Unlike SLAM-based methods, our approach does not require any camera calibrations, such as pre-calibrated extrinsic and focal.
\subsection{Camera System and Camera Pose Initialization}
\label{subsec_pose}
\textbf{Initialization:} To ensure inter-camera consistency, we first identify a central camera and align all other cameras to it. Given the first frame of all cameras $\{I_k^0\}_{k=0}^K$, we first perform fast feature extraction using \cite{potje2024xfeat}, followed by pairwise feature matching and computation of pixel-wise feature distances, as shown in Fig.~\ref{fig_framework} (a). The same strategy is also adopted for keyframe selection (see Appendix for details). The central camera is determined as the one minimizing the sum of pairwise distances to all other cameras. For more complex camera rigs, we first filter out valid matching groups and then hierarchically construct a camera tree, where the root node of each layer corresponds to the central camera. The cameras are subsequently aligned layer-by-layer. Once the central camera \( m \) is determined, we une the first \( N_{\text{init}} \) frames $\{I_m^t\}_{t=0}^{N_{init}}$ of central camera to perform exhaustive feature matching between all pairs of these frames and jointly optimize the focal length, camera poses, and 3D keypoint positions by minimizing the reprojection error with bundle adjustment (BA). Subsequently, we match the \( N_{\text{init}} \) frames of each remaining camera with those of the central camera. For each pair, as shown in Fig.~\ref{fig_framework} (b), we select the subset of correspondences associated with already registered 3D points using GPU-parallel RANSAC and perform a mini BA to estimate the relative transformation matrix $\hat{\mathbf{T}}_{m\rightarrow{} k}^t \in SE(3) $  and align both inter-camera depths and 3D points. The transformation $\hat{\mathbf{T}}_{m\rightarrow{}k}^{t^*}$ with the lowest reprojection error is adopted as the initial relative transformation, which is further refined in the subsequent optimization stage.\\
\textbf{Subsequent Pose Estimation:}
In each new frame at time \( t \), we still estimate the pose of every camera by performing feature matching followed by BA. Specifically, for each camera \( i \in \{0, \dots, K\} \), we extract keypoints and establish correspondences with the previously registered 3D points. The objective is defined as:

\begin{equation}
\underset{\{\mathbf{T}_{i}^{t}\}_{i=1}^{K}}{\arg\min} 
\sum_{i=1}^{K} \sum_{j \in \mathcal{P}} 
\left\| 
\pi\!\left( \mathbf{T}_{i}^{t} \mathbf{X}_{j} \right) - \mathbf{x}_{i,j}^{t} 
\right\|_{2}^{2},
\label{eq:online_ba_argmin}
\end{equation}
Where \( \mathbf{T}_{i}^{t} \) is camera pose for camera i and time $t$ to be optimized, \( \mathbf{X}_{j} \) denotes the 3D position of point \( j \), 
\( \mathbf{x}_{i,j}^{t} \) is its 2D observation in camera \( i \) at time \( t \), $\mathcal{P}$ is the matched 3D points,
and \( \pi(\cdot) \) is the perspective projection function. However, jointly optimizing the poses of all cameras is highly inefficient and prone to drift solutions, especially under wide-baseline. To improve efficiency while maintaining stable trajectories of the camera rig, 
we maintain a dedicated 3D point subset for each camera and establish 3D-2D correspondences only within its own subset, which results in a fixed-size sparse Jacobian of the reconstruction error, accelerates the correspondence estimation and BA process and reduces cross-camera mismatches. Considering the rigid linkage of the camera rig, the extrinsics of camera \( i \) at time \( t \) can be derived as
$\mathbf{T}_{i}^{t} = \mathbf{T}_{m}^{t} \mathbf{T}_{i \rightarrow{} m}$,
where \( \mathbf{T}_{i \rightarrow{} m} \) is the relative transformation between camera \( i \) and the central camera. The objective in Eq.~\ref{eq:online_ba_argmin} can be written as:
\begin{equation}
\underset{\mathbf{T}_{m}^{t}}{\arg\min} 
\sum_{i=1}^{K} \sum_{j \in \mathcal{P}_i} 
\left\| 
\pi\!\left( \mathbf{T}_{m}^{t} \mathbf{T}_{i \rightarrow{} m} \mathbf{x}_{j} \right) - \mathbf{x}_{i,j}^{t} 
\right\|_{2}^{2},
\label{eq:online_ba_argmin_2}
\end{equation}
As a result, we only need to estimate the optimal transformation $ \mathbf{T}_{m}^{t}$ for the central camera to obtain the optimal pose for all cameras at the current time step, which significantly simplifies the optimization process and improves the trajectory stability. To further accelerate optimization, we perform gradient propagation on a randomly selected camera’s subset points in each iteration, while computing the reprojection error across all cameras to seek the optimal solution. We also introduce an inter-camera balancing strategy to prevent the optimization from falling into local minima, details are provided in appendix.

\subsection{Redundancy-free Gaussian Sampling}
\label{subsec_Gaussian}
Since the densification process of Gaussian primitives is highly inefficient, most on-the-fly reconstruction methods directly sample from the input images for Gaussian primitives initialization and disable densification during subsequent optimization. However, due to substantial view overlaps both between consecutive frames of the same camera and across different cameras within the same timestamp, naïvely sampling every pixel into a Gaussian primitive introduces severe spatial and temporal redundancy. \textbf{To mitigate inter-frame redundancy,} 
we prioritize high-frequency regions and poorly reconstructed areas by computing an insertion probability at each pixel $(u, v)$ 
using the Laplacian of Gaussian (LoG) operator same as~\cite{meuleman2025on,li2025artdeco}:

\begin{equation}
\begin{aligned}
P_a(u, v) = 
\max\Big(
(\mathcal{P}(I)(u,v)  - \mathcal{P}(\tilde{I})(u,v)),0\Big)
\end{aligned}
\label{eq:insert_prob}
\end{equation}
where:
\begin{equation}
\begin{aligned}
\mathcal{P}(I)(u,v) = \min\big(\|\nabla^2(G_\sigma) * I(u, v)\|,\, 1\big)
\end{aligned}
\label{eq:insert_prob2}
\end{equation}
where $I$ and $\tilde{I}$ are the ground-truth and rendered images, and $G_\sigma$ is a Gaussian kernel with standard deviation $\sigma$. 
A new Gaussian is added when $P_a(u, v)$ exceeds the threshold $\tau_a$.
the corresponding primitives scale scale at pixel $(u,v)$ is defined as:

\begin{equation}
S = \frac{d}{2\sqrt{\mathcal{P}(I)(u,v)}f}, 
\label{eq:base_scale}
\end{equation}
where $d$ is the distance from the Gaussian center to the camera computed by depth and $f$ is the focal length. \textbf{To address inter-camera redundancy}, we adopt a per-camera Gaussian merging strategy. 
Assume that a set of Gaussian primitives has already been established from cameras $0$ to $i\!-\!1$. 
When adding new Gaussians based on camera $i$, we first reproject the existing Gaussian primitives 
onto the image plane of camera $i$, and compute their corresponding 2D coordinates $x'$, depths $d'_i$. We then use bilinear interpolation to match the reprojected pixel $\mathbf{x}'_i$ and the sampled pixel $\mathbf{x}_i$ by Eq.~\ref{eq:insert_prob2}.
All existing Gaussian primitives without pixel correspondences are retained for further optimization. 
For paired primitives, we first compute their respective image-space scales:
\begin{equation}
s' = 
\frac{2Sf}{
d'_i
},
\label{eq:img_scale}
\end{equation}
where $d'_i$ is the distance from the existing Gaussian center to camera i.
\begin{equation}
s_i = 
\frac{1}{
2 \sqrt{\min(\|\nabla^2 G_\sigma * I_i(u, v)\|, 1)}
},
\label{eq:img_scale}
\end{equation}
Here, $s'$ denotes the reprojected image-space scale of existing Gaussian primitives obtained from cameras $0$ to $i-1$,
while $s$ represents the image-space scale initialized from camera $i$ using the LoG-based sampling.
Subsequently, we perform Gaussian merging based on the image-space scale.
We filter out primitives with larger image-space scales and retain those with smaller scales,
preserving all their attributes except color, which is interpolated between the two or more primitives, as shown in Fig.~\ref{fig_framework}.

During this stage, depth values are also re-aligned by those merged Gaussian primitives for consistency.
Note that pixel-corresponding Gaussian primitives with significant depth discrepancies ($\|d'_i-d_i\|>\tau_d$) are not merged, to prevent fusing Gaussian primitives from different surfaces and to preserve occlusion relationships.
\subsection{Optimization and Scheduling}
\label{subsec_optimize}
\textbf{Optimization:} To capture low-frequency scene details first and accelerate optimization. We employ a linear level-of-detail (LOD) structure for coarse-to-fine training and adaptively increase the downsampling level based on the DFT, which is discussed in the following paragraph. During optimization, both the Gaussian primitive parameters and camera parameters are jointly updated. 
When updating camera parameters, gradients from the spherical harmonics are detached. 
For non-central cameras, we optimize the relative transformation matrices with respect to the central camera. Same as \cite{meuleman2025fly}, we adopt an active anchor-building strategy for aggregating distant Gaussian points to reduce computational and memory load. Moreover, we design a dedicated mechanism for multi-camera rigs to detect when to create new anchors. The detailed description of this component is provided in the appendix.\\
\textbf{Frequency-based Scheduling:} Due to the incremental scene construction and LOD design, our optimization naturally progresses from focusing on low-frequency structures to refining high-frequency details. This progress, however, varies across regions. Areas with denser Gaussian initialization or richer high-frequency details require a slower transition and more iterations, while others converge with fewer updates.  To achieve on-the-fly reconstruction, the number of optimization iterations per keyframe is strictly limited (typically 10–30).
it is crucial to allocate optimization steps adaptively, focusing more on areas (keyframes) that demand finer refinement. We design a frequency-based scheduler that allocates more optimization steps to regions still dominated by low-frequency components. For each frame, we compute:
\begin{equation}
r = 
\frac{\mathcal{X}(\mathbf{I})}{\mathcal{X}(\tilde{\mathbf{I}})}
\label{eq:freq_fraction}
\end{equation}
where
\begin{equation}
\mathcal{X}(\mathbf{I}) = 
\sum_{i=1}^{h}
\sum_{j=1}^{w}
\|{\mathrm{DFT}(\mathbf{I})(i,j)\|_2}
\label{eq:freq_significance}
\end{equation}
and
\begin{equation}
\mathcal{X}(\mathbf{\tilde{I}}) = 
\sum_{i=1}^{h}
\sum_{j=1}^{w}
\|{\mathrm{DFT}(\mathbf{\tilde{I}})(i,j)\|_2}
\label{eq:freq_significance_render}
\end{equation}
where $I$ and $\tilde{I}$ are the ground-truth and rendered images and $\mathrm{DFT}$ denotes the discrete Fourier transform. Before each optimization iteration, we sample frames to be optimized according to their frequency-based weights $r$. Meanwhile, we also determine whether the LOD structure requires upsampling based on the updated frequency ratio $r'$ where:
\begin{equation}
r' = 
\frac{\mathcal{X}(\mathbf{I}^n)}{\mathcal{X}(\tilde{\mathbf{I}})}
\label{eq:freq_fraction}
\end{equation}
Here, $n$ denotes the downsampling factor. If the updated ratio $r'$ is below $\tau_f$, we increase the ground-truth resolution by one level.
\section{RigScapes Datasets}
\begin{figure}[ht]
\vspace{-0.1cm}
  \centering
  \includegraphics[width=0.8\columnwidth]{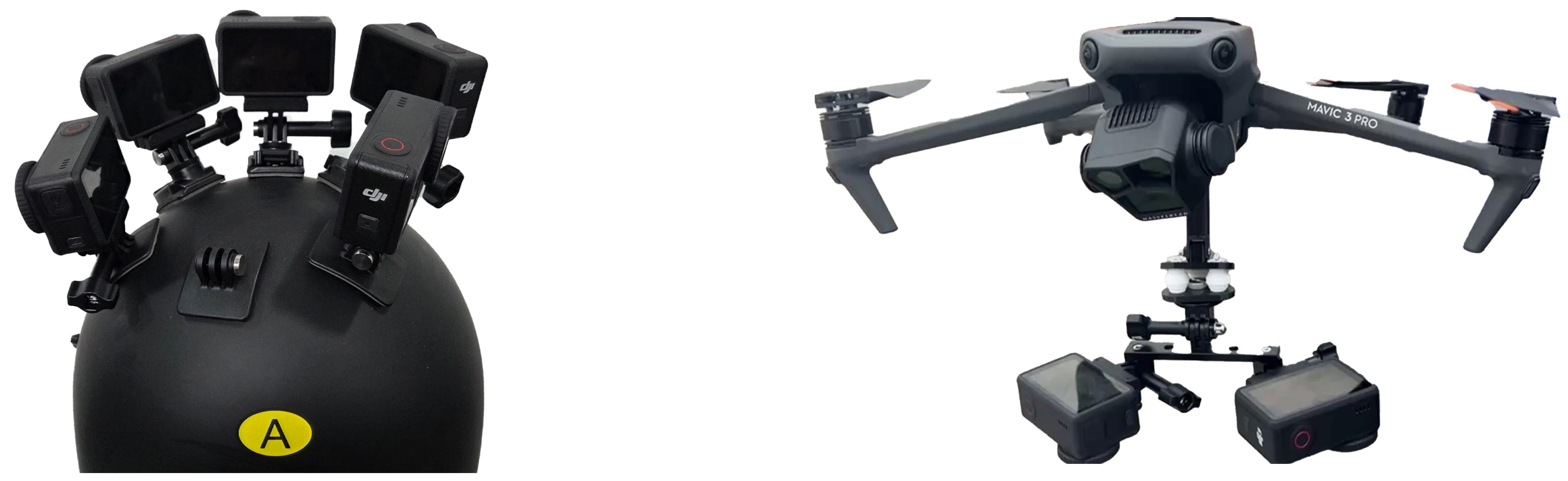}
  \vspace{-0.2cm}
  \caption{\textbf{Left:} Our 5 DJI Action 5 Pro camera helmet rig. \textbf{Right:} Our two synchronized DJI Action 5 Pro on
DJI Mavic 3 Pro drone. }
  \label{fig_dataset}
  \vspace{-0.3cm}
\end{figure}

Many existing multi-camera rig datasets suffer from lens distortion, inconsistent exposure, homogeneous scene types, excessive dynamic objects, and limited scene scales, making them unsuitable for evaluating fast 3D reconstruction algorithms. To address these issues, we introduce the RigScapes dataset, which contains over 1 km of road scenes and 1 km² of aerial scenes, all captured using consumer-grade cameras (DJI Action 5 Pro).
As shown in Fig.~\ref{fig_dataset}, we used a bicycle helmet on which
we mounted 5 cameras, set to Linear FoV and 1920×1080
resolution with 4 FPS for road captures. 
For aerial captures, we mount two synchronized cameras on a DJI Mavic 3 Pro drone to capture the city, set to Linear FoV and 1920×1080 resolution with 2 FPS. We carefully maintain consistent exposure between cameras and avoid capturing excessive dynamic objects. All cameras are synchronized by Osmo Action GPS bluetooth remote controller. We will also provide both images and videos to facilitate the evaluation of reconstruction latency. More details are provided in the appendix.
\section{Experiments}
\begin{table*}[htb]
\vspace{-0.3cm}
\setlength{\tabcolsep}{0.6mm}
\centering
\small
\begin{tabular}{c|c|c|cccc|cccc|cccc|}
\hline 
\multirow{2}{*}{\textbf{Input}} & \textbf{Datasets}&\multirow{2}{*}{\textbf{calib.}} & \multicolumn{4}{c|}{SmallCity} & \multicolumn{4}{c|}{MatrixCity} & \multicolumn{4}{c|}{RigScapes} \\
 \cline{2-2} \cline{4-15}
& \textbf{Methods$\vert$Metrics}& & PSNR$\uparrow$ & SSIM$\uparrow$ & LPIPS$\downarrow$ & Time$\downarrow$ 
& PSNR$\uparrow$ & SSIM$\uparrow$ & LPIPS$\downarrow$ & Time$\downarrow$ 
& PSNR$\uparrow$ & SSIM$\uparrow$ & LPIPS$\downarrow$ & Time$\downarrow$\\
\hline

\multirow{6}{*}{\rotatebox{90}{Mono}} 
& MonoGS &\ding{51}  &10.79 &0.380 &0.620 & 00:07:11. &14.21 &0.339 &0.592 &00:13:41 &10.66 &0.281 &0.659 &00:09:18\\
& WildGS-SLAM &\ding{51}  &10.35 &0.290 &0.637 &01:27:11 &16.03 &0.385 &0.604 &00:17:43 &7.93 &0.139 &0.785 &00:57:19\\
& Longsplat&\ding{55}  &9.76 &0.230 &0.603 &04:44:42 &16.34 &0.323 &0.523 &00:27:25 &8.99 &0.144 &0.670 &00:04:11 \\
& HT-3DGS &\ding{55}  &13.99 &0.428 &0.523 &03:25:53 &16.90 &0.481 &0.612 &00:49:12 &15.33 &0.379 &0.427 &08:55:51\\
& On-the-fly-NVS&\ding{55}   &15.87 &0.517 &0.417 \cellcolor{pink!70} &00:01:18 &18.02 &0.512 &0.426 &00:00:39 &14.82 &0.424 &0.531 &00:03:11\\
\hline
\multirow{7}{*}{\rotatebox{90}{Multi}} 
& MonoGS& \ding{51}  &10.54 &0.295 &0.798 &00:19:43 &14.47 &0.347 &0.743 &00:13:19 &11.65 &0.296 &0.734 &00:18:16\\
& WildGS-SLAM&\ding{51} & 9.51 & 0.303&0.821 &04:12:45 & 16.09& 0.314& 0.802& 00:57:21& 11.10&0.285 &0.723 &06:56:23\\
& Longsplat&\ding{55}   &14.61 &0.426 &0.548 &17:20:37 &20.64 &0.514 &0.401 &00:50:07 &11.28 &0.185 & 0.626 &13:58:47 \\
& HT-3DGS& \ding{55}  &11.53 &0.367 &0.661 &11:15:12 &22.35 &0.600 &0.538 &02:25:20 &11.30 &0.265 &0.694 & 08:11:53\\
& On-the-fly-NVS& \ding{55}  &17.88 \cellcolor{yellow!40} &0.587 \cellcolor{yellow!40} &0.486 &00:04:53 \cellcolor{yellow!40} &28.18 \cellcolor{yellow!40} &0.819 \cellcolor{yellow!40} &0.258 \cellcolor{yellow!40} &00:01:09 \cellcolor{pink!70} &16.45 \cellcolor{yellow!40} &0.445 \ &0.527 \cellcolor{yellow!40}  &00:08:15 \cellcolor{yellow!40}\\
& \textbf{Ours}& \ding{55} &21.35 \cellcolor{pink!70} & 0.659\cellcolor{pink!70} & 0.420\cellcolor{yellow!40} &00:02:12 \cellcolor{pink!70} 
&29.61 \cellcolor{pink!70} &0.867 \cellcolor{pink!70} &0.220 \cellcolor{pink!70} &00:01:30 \cellcolor{yellow!40} 
&20.12 \cellcolor{pink!70} &0.602 \cellcolor{pink!70} &0.386 \cellcolor{pink!70} &00:04:25 \cellcolor{pink!70} \\
\hline

\end{tabular}
\caption{ Quantitative comparisons of novel view synthesis and reconstruction time. \textbf{Best} results are marked in \colorbox{pink!70}{red} and the second best results are marked in \colorbox{yellow!40}{yellow}. The Calib. column indicates whether camera calibration is required. We provide both mono and multi-camera input modes for all baselines. Runtime is compared only among methods that use multi-camera inputs.}
\vspace{-0.3cm}
  \label{tab_main}
\end{table*}
\begin{figure*}[t]
  \centering
  \includegraphics[width=0.96\linewidth]{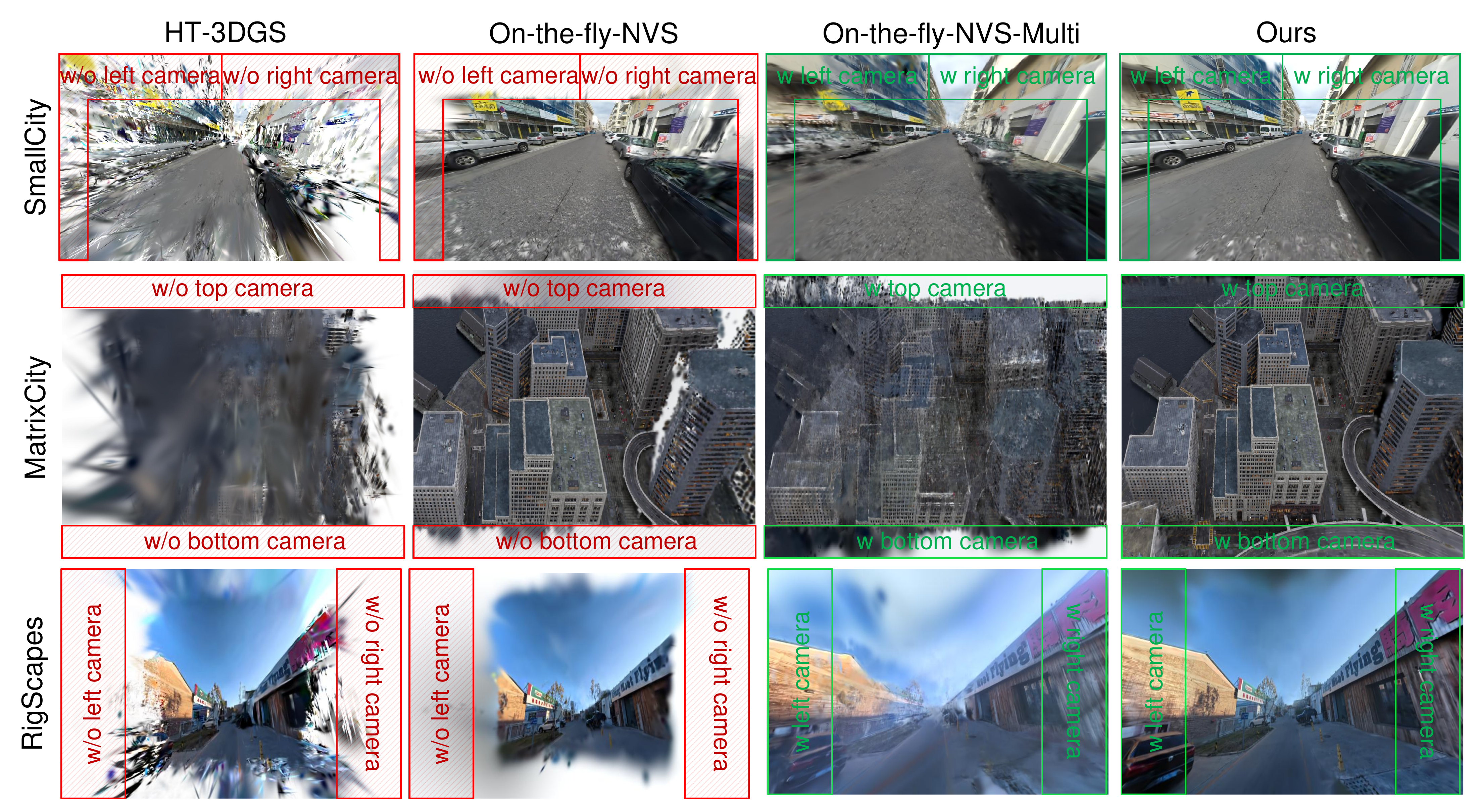}
  \caption{Visualizations with the changed FOV. Using monocular video streams often leads to severe scene incompleteness.
Directly applying monocular methods to multi-camera streams results in significant scene inconsistency and noticeable artifacts.
In contrast, our approach achieves complete and high-fidelity scene reconstruction.}
  \label{fig_fov}
  \vspace{-0.5cm}
\end{figure*}
\subsection{Experiment Settings}
\textbf{Datasets.} We conduct extensive experiments on three challenging large-scale datasets with varying scenes, camera settings and conditions.
\textbf{H3DGS SmallCity Dataset (Road) \cite{kerbl2024hierarchical}}: A 450 m long road captured by Gopro rigs. We use two camera configurations: Cam0–1–2 (easy, with large inter-camera overlap) Cam0–2–4 (hard, with small inter-camera overlap), to evaluate our method and all baselines' robustness under different camera settings. Since all baseline methods suffer from out-of-memory (OOM) errors or severe trajectory drift when using data from more than three cameras, we limit the comparison experiments to at most three camera inputs.
\textbf{MatrixCity Dataset (Aerial) \cite{li2023matrixcity}}: We selected aerial data from four blocks of Matrix City, with block areas ranging from 0.15 km² to 0.5 km². Consistent with the previous experiments, we still use three cameras as input.
\textbf{Rigscapes Dataset (Aerial+Road)}: A dataset that includes a 1 km road-scan sequence and a 1 km² aerial area. Detailed descriptions can be found in the dataset section. \\
\noindent{\textbf{Metrics.}} For novel view synthesis, we employ quantitative metrics including PSNR, SSIM, and LPIPS for synthetic data. Regarding camera pose estimation, we quantify performance using both absolute and relative error metrics, specifically reporting ATE and RPE. \\
\textbf{Baselines.} We first compare our method with SLAM-based methods: MonoGS \cite{matsuki2024gaussian} and WildGS-SLAM \cite{zheng2025wildgs}. Then, we compare our method against off-line pose-free methods, such as HT-3DGS \cite{ji2025sfm} and Longsplat \cite{lin2025longsplat}. Finally, we benchmark our approach against the previous SOTA large-scale on-the-fly 3DGS methods: On-the-fly-NVS \cite{meuleman2025fly}. Note that we design two evaluation modes for the baseline methods: a monocular mode, which uses only the center camera (mono in Tab.~\ref{tab_main}), and a sequential multi-camera mode (multi in Tab.~\ref{tab_main}), which processes multi-camera streams as a temporal sequence. 
\textbf{In addition to using a fixed test hold, we also utilize additional trajectories—featuring deliberate rotations and translations relative to the (central) camera—to evaluate each method’s novel view synthesis capability under more challenging conditions.}
For pose estimation, we additionally compare with feed-forward methods: Fast3R \cite{yang2025fast3r} and $\pi^3$ \cite{wang2025pi}.

\begin{figure*}[!h]
  \centering
  \vspace{-0.5cm}
  \includegraphics[width=0.98\linewidth]{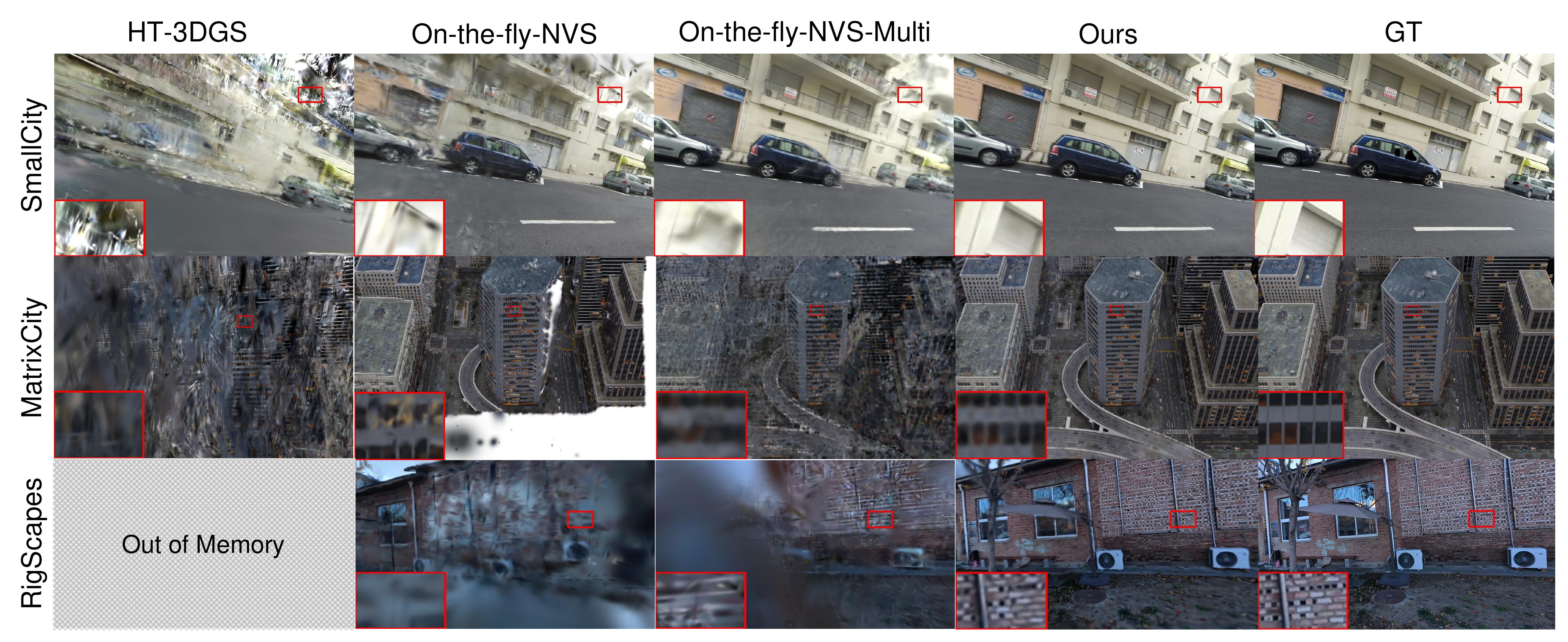}
  \caption{Visualization results after camera rotation. 
Only our method produces visually satisfactory results under such extreme novel-view settings.}
  \label{fig_align}
  \vspace{-0.5cm}
\end{figure*}

\noindent\textbf{Implementation Details.} All experiments were conducted on NVIDIA A100 GPU. For hyperparameters, we set $N_{init}=8$, $\tau_a=0.2$ and $\tau_f=2.0$. For road scenes, we set optimization iterations to 10 per camera, whereas for aerial drone scenes, the iteration count is increased to 60 to account for the slower and more gradual camera motion.

\subsection{ Experimental Results}
\textbf{Novel View synthesis Results:}
We report the comparison results of novel view synthesis tasks in Tab.~\ref{tab_main}. Our method consistently outperforms the others across all metrics. On average, our method achieves \textbf{a 2.82dB higher PSNR and 15$\textbf{\%}$ higher SSIM} compared to the best performance among all baselines. Due to occasional OOM failures for certain methods on specific scenes, we report the mean metrics computed over valid reconstructions only. A per-scene breakdown is provided in the appendix.
In Fig.~\ref{fig_fov}, we present the rendering results under the test-time view with an enlarged field of view (FOV). Compared with using only monocular input, our method clearly achieves a more complete reconstruction, benefiting from the richer scene information provided by the multi-camera rig. In contrast to other methods that also take multi-camera inputs, our approach produces more consistent, high-quality, and sharper results without noticeable artifacts, further demonstrating its advantage in handling multi-camera rigs. 
In Fig.~\ref{fig_align}, we show novel view synthesis results from extreme viewpoints that lie outside the main camera trajectory. As observed, all baseline methods fail under such challenging views, producing incomplete geometry or noticeable artifacts and inconsistencies. In contrast, our method maintains exceptionally high rendering quality, even under these extreme viewpoint conditions. 
All evaluations are conducted on large-scale scenes (over 100 m or 0.1 km²). \textbf{We provide supplementary video for further reference.}\\
\begin{figure}[t]
\vspace{-0.3cm}
  \centering
  \includegraphics[width=1.05\columnwidth]{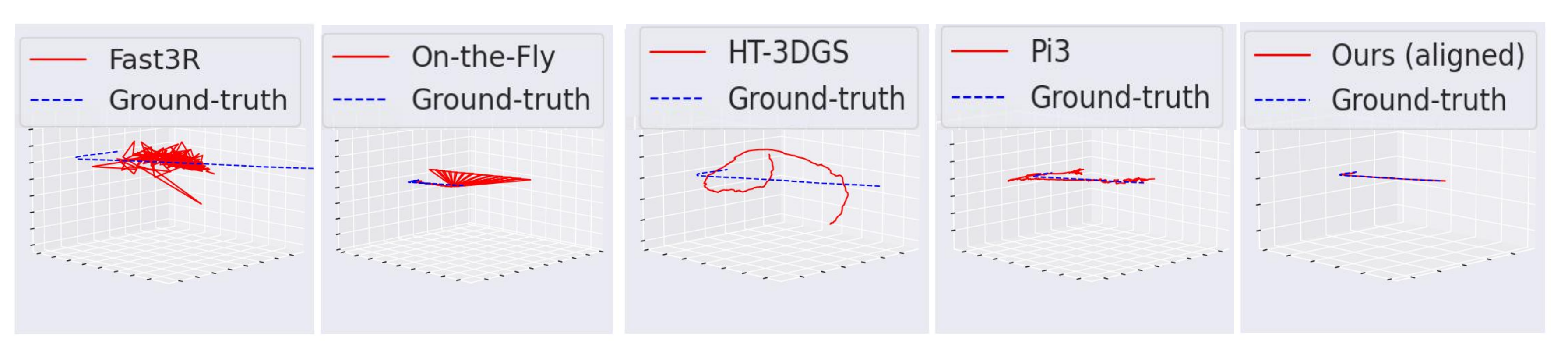}
  \vspace{-0.4cm}
  \caption{Visualization of multi-camera rig trajectories.}
  \vspace{-0.4cm}
  \label{fig_pose}
\end{figure}
\noindent{}\textbf{Pose Estimition Results:}
The quantitative results for camera pose accuracy on SmallCity dataset are presented in Tab.~\ref{tab_pose}. The learned camera poses of all methods are post-processed and aligned with ground truth using a consistent method following \cite{ji2025sfm}. The results show that the camera pose accuracy of our method consistently not only outperforms previous pose-free 3DGS methods but also is superior to Feed-forward methods. In Fig.~\ref{fig_pose}, we further visualize the estimated camera trajectories. As shown, all previous methods suffer from severe drift or even complete failure when estimating camera poses for wide-baseline multi-camera rigs. In contrast, our method produces highly accurate and stable trajectories, thanks to our carefully designed camera system initialization, alignment, and lightweight multi-camera bundle adjustment (BA) strategy. We further analyze this aspect in our ablation studies.\\
\begin{table}[ht]
\centering
\small
\vspace{-0.2cm}
\begin{tabular}{l|ccc}
\hline  
Method & $\text{RPE}_t \downarrow$ & $\text{RPE}_r \downarrow$ & $\text{ATE} \downarrow$ \\
\hline  
HT-3DGS & 0.171  &5.164  &0.035  \\
Pi3 & 0.341  &5.208  &0.015  \\
Fast3R & 1.080  &42.139  &0.062  \\
On-the-fly-NVS &0.253  &3.742  & 0.009 \\
Ours & \textbf{0.075} & \textbf{1.991} & \textbf{0.005} \\
\hline  
\end{tabular}
\vspace{-0.2cm}
\caption{ Quantitative comparisons of pose estimation.}
\vspace{-0.4cm}
  \label{tab_pose}
\end{table}
\begin{table}[ht]
\centering
\begin{tabular}{l|ccc}
\hline  
module& Time (ms) \\
\hline  
Data Loading\& Feature Extraction& 9.1 \\
PnP+RANSAC \& Multi-Camera BA & 28.6\\
Reducy-Free Gaussian Sampling &33.7\\
Optimization & 113.3\\
Scheduling  &  0.3 \\
\textbf{Per Keyframe} & 185 \\
\textbf{Per Keyframe 3cam} & 555\\
\hline 
\end{tabular}
\caption{Per keyframe runtime breakdown.}
  \label{tab_time}
\end{table}

\noindent{}\textbf{Efficiency:} In Tab.~\ref{tab_main}, we report the reconstruction efficiency of our method compared to baseline approaches. On average, our method can reconstruct approximately 100 meters of road or 100,000 m² of aerial scenes within two minutes, making it about 300× faster than offline methods. Under wide-baseline and multi-camera input settings, our approach also shows significantly higher efficiency than SLAM-based and other online methods.
In Tab.~\ref{tab_time}, we further detail the efficiency of each module. Since the runtime of several components depends on the number of cameras, we report per-camera efficiency. Our method can process a per-camera
keyframe in around 200 ms, which is sufficient for on-the-fly reconstruction.

\subsection{Ablation Studies}
\begin{figure}[t]
\vspace{-0.2cm}
  \centering
  \includegraphics[width=1.05\columnwidth]{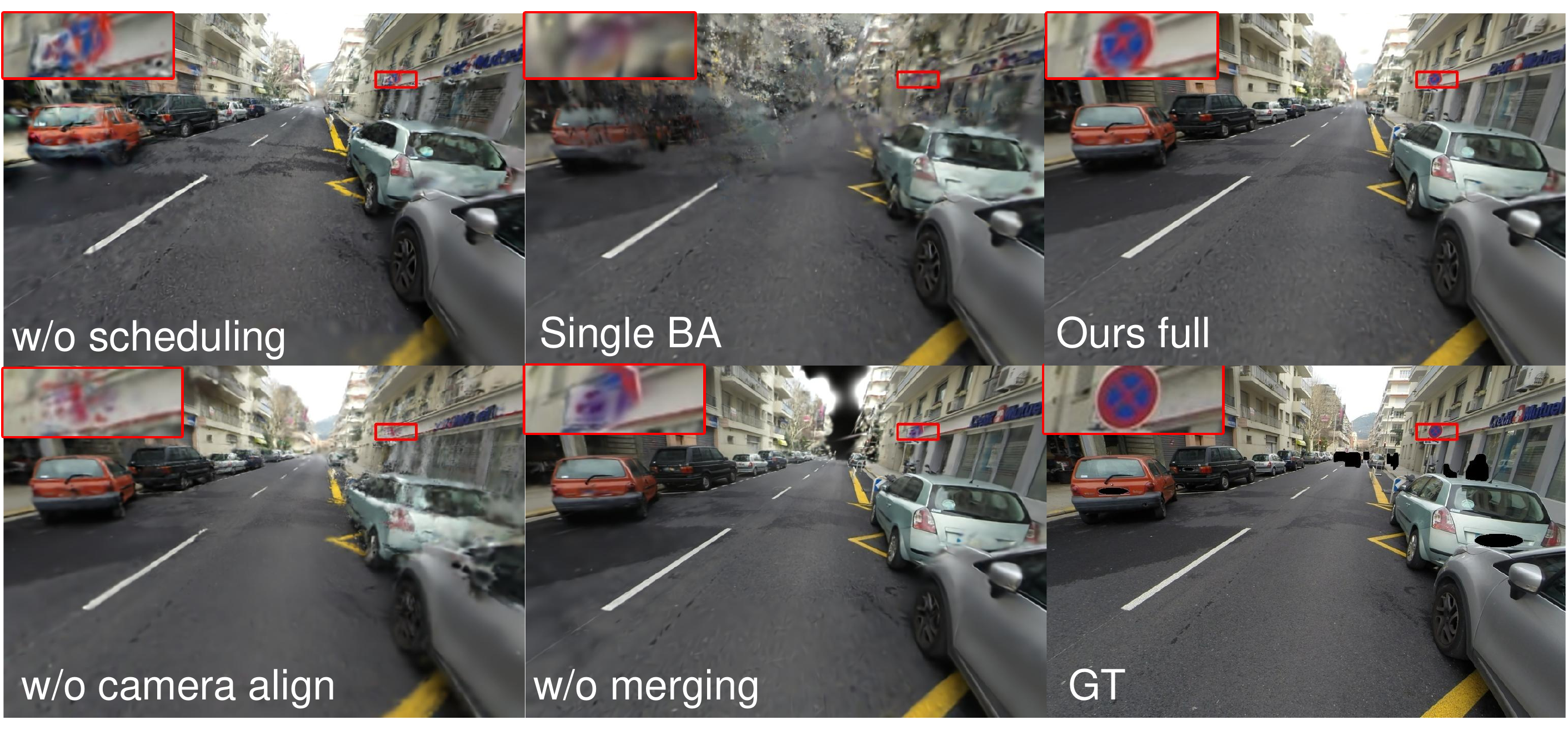}
  \vspace{-0.4cm}
  \caption{Visualization results for ablation study.}
  \label{fig_ab}
   \vspace{-0.4cm}
\end{figure}
\begin{figure}[t]
\vspace{-0.0cm}
  \centering
  \includegraphics[width=1.05\columnwidth]{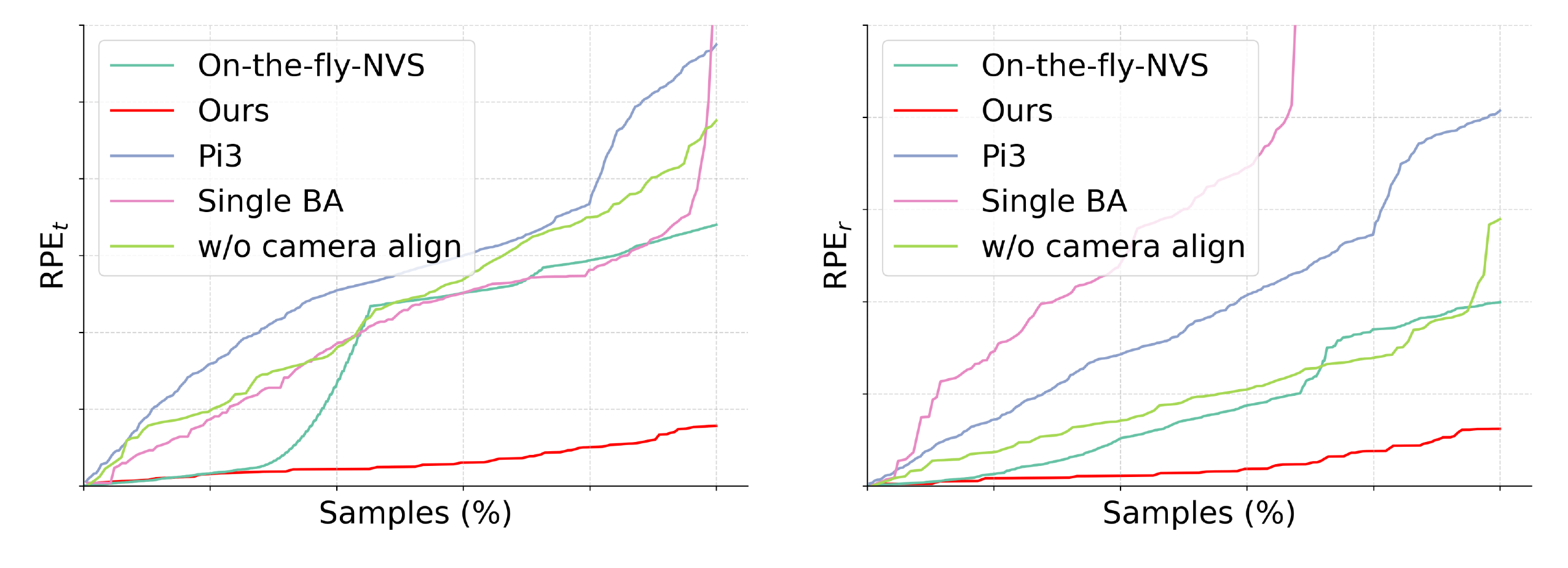}
  \vspace{-0.3cm}
  \caption{Visualization results for ablation study.}
  \vspace{-0.4cm}
  \label{fig:pose_ab}
\end{figure}

We perform ablation studies to illustrate the importance of each design. First, we replace our camera system construction and center-alignment scheme with a direct pose and scene initialization using all cameras (w/o camera align).
Second, we substitute our multi-camera BA strategy with independent single-camera BA for each camera (single BA).
Third, we disable our frequency-based scheduler (w/o scheduling).
Finally, we disable Gaussian merging (w/o merging). We evaluate the impact of these design choices on rendering quality, camera trajectory estimation, and memory usage using the SmallCity dataset.\\
\begin{table}[!h]
\centering
\begin{tabular}{l|ccc}
\hline  

Method & PSNR$ \uparrow$ & SSIM$\uparrow$ & LPIPS$\downarrow$ \\
\hline  
Single BA &  12.67  &0.421  &0.657  \\
w/o camera align &  16.07  &0.515  &0.561  \\
w/o scheduling& 18.02  &0.563  &0.471  \\
w/o merging &19.92  &0.632  &  0.451\\
Ours full& \textbf{21.35} & \textbf{ 0.659} & \textbf{0.420} \\
\hline  
\end{tabular}
\caption{ Quantitative ablation on the SmallCity dataset. }
  \label{tab_ab}
\end{table}
\nocite{}\textbf{Rendering Quality:} In Tab.~\ref{tab_ab}, we present the novel view synthesis results obtained after selectively disabling individual components, revealing that each proposed strategy contributes substantially to the overall rendering quality. Corresponding visual comparisons are provided in Fig.~\ref{fig_ab}. As illustrated, disabling multi-camera bundle adjustment or camera alignment results in pronounced artifacts stemming from trajectory drift and cross-camera inconsistency. The absence of frequency-based scheduling leads to noticeable degradation in local detail, suggesting insufficient optimization in under-constrained regions. Moreover, disabling Gaussian merging introduces structural artifacts and voids, as the lack of depth-consistent alignment causes unstable addition and removal of local geometric details.\\
\noindent{}\textbf{Trajectory Estimation:} 
We further evaluate robustness through the cumulative RPE error distributions (translation and rotation) in Fig.~\ref{fig:pose_ab}
Our method consistently achieves lower errors, ensuring stable trajectories.
In contrast, using single-camera BA or removing camera alignment leads to severe drift and instability, confirming the effectiveness of multi-camera BA and alignment strategy.\\
\noindent{}\textbf{Memory Usage:}
In Tab.~\ref{tab_mem}, we report the peak Gaussian count, average Gaussian count, and average memory consumption per optimization cycle, with and without the Gaussian merging strategy. As observed, all three metrics are substantially reduced when merging is enabled, demonstrating that the Gaussian merging strategy effectively mitigates inter-camera redundancy. 
\begin{table}[ht]
\centering
\small
\begin{tabular}{l|ccc}
\hline  
& AVG GS $\downarrow$ & Peak GS $\downarrow$ &Memory $ \downarrow$ \\
\hline  
w/o Merging & 674278 &855426  &  6374MB \\
w Merging& 585672 & 788137 &  5699MB \\
\hline 
\end{tabular}
\caption{Impact of the Gaussian merging on SmallCity. }
\vspace{-0.3cm}
\label{tab_mem}
\end{table}

\section{Conclusion}
We introduced the first on-the-fly 3D reconstruction framework tailored for large-scale multi-camera rigs, enabling kilometer-scale scene reconstruction within minutes and without explicit calibration. Through our centralized initialization, lightweight multi-camera BA, redundancy-free Gaussian sampling, and frequency-based optimization, the proposed system achieves both high efficiency and fidelity across wide baselines.
{
    \small
    \bibliographystyle{ieeenat_fullname}
    \bibliography{main}
}


\end{document}